\documentclass{article}
\usepackage[table]{xcolor}
\usepackage{nips15submit_e,times}
\usepackage{url}
\usepackage{listings}
\usepackage{graphicx}
\usepackage[]{algorithm2e}
\usepackage[utf8]{inputenc}

\usepackage{amssymb}
\usepackage{pifont}

\usepackage[cmex10]{amsmath}

\title{Towards a more efficient representation of imputation operators in TPOT}
\author{
	Unai Garciarena\thanks{\texttt{unai.garciarena@ehu.es}} \\
	Intelligent Systems Group  \\
	Univ. of the Basque Country\\
	(UPV/EHU) \\
	San Sebastian, Spain \\
		\And
		Alexander Mendiburu \thanks{\texttt{alexander.mendiburu@ehu.es}}\\
		Intelligent Systems Group \\
		Univ. of the Basque Country\\
		(UPV/EHU) \\
		San Sebastian, Spain 
	\And
	Roberto Santana\thanks{ \texttt{roberto.santana@ehu.es}}  \\
	Intelligent Systems Group \\
	Univ. of the Basque Country\\
	(UPV/EHU) \\
	San Sebastian, Spain \\
}

\nipsfinalcopy 

\begin{document}

	\maketitle
	
	\begin{abstract}
		Automated Machine Learning encompasses a set of meta-algorithms intended to design and apply machine learning techniques (e.g.,  model selection, hyperparameter tuning,  model assessment, etc.). TPOT, a software for optimizing machine learning pipelines based on genetic programming (GP), is a novel example of this kind of applications. Recently we have proposed a way to introduce imputation methods as part of TPOT. While our approach was able to deal with problems with missing data, it can produce a high number of unfeasible pipelines. In this paper we propose a strongly-typed-GP based approach that enforces constraint satisfaction by GP solutions. The enhancement we introduce is based on the redefinition of the operators and implicit enforcement of constraints in the generation of the GP trees. We evaluate the method to introduce imputation methods as part of TPOT. We show that the method can notably increase the efficiency of the GP search for optimal pipelines.  \\
		
		{\bf{keywords}}: genetic programming, missing data, imputation methods, supervised classification, automatic machine learning
	\end{abstract}

	\section{Introduction}
	Tree-based Pipeline Optimization Tool (TPOT) is a software that allows its user to find combinations of machine learning (ML) algorithms to build a model out of raw data.  These combinations of processes, named pipelines in this context, are evolved using genetic techniques, more specifically, genetic programming. The evolution produces improvements in both the algorithm selection, and parameter tuning, checking a vast amount of variants, which leads to a very complete analysis of the interactions of all the components involved. Even though the initial versions did not support incomplete data, i.e., datasets with missing data (MD), more actual releases do. However, the way TPOT deals with MD is by applying a simple imputation method (IM) at the start of the algorithm, then performing its regular pre-precessing and classification/regression routines. This approach completely excludes imputation from the evolution, and, as it has been shown in several works, the interactions between IMs and supervised classification algorithms can be strong, and should be taken into account for achieving higher classification accuracies in some cases \cite{garciarena_extensive_2017,luengo_study_2010}. This is the reason why, in a previous work, we have proposed a method for introducing IMs as additional components of the pipelines evolved by TPOT \cite{garciarena_evolving_2017}. However, our previous proposal exhibited one important limitation; it produced a large amount of infeasible pipelines, and therefore its results were probably too biased towards the initial random generation of pipelines. This behavior produced that the evolutionary effort of the software was barely exploited.
    
    This paper is a continuation of \cite{garciarena_evolving_2017}. The contribution described here is the introduction of a more sophisticated approach. However, the context of applications and much of the relevant work is similar to that presented in \cite{garciarena_evolving_2017}. Therefore, we recommend reading \cite{garciarena_evolving_2017} in order to understand the software taken as a baseline for this contribution, the details of the enhancement, and the conclusions drawn from the experimentation. From now on, this work assumes previous knowledge on all those topics.
    
    In our previous proposal, the IMs were added as simple preprocessing units (such as those already available in TPOT, e.g., PCA, Feature extraction, ...). This resulted in populations with high amount of invalid individuals, as only pipelines that contained an IM in the initial position could be successfully evaluated. Even though the search space kept being the same, the capacity of the algorithm of searching locations far from those in which the valid initial individuals were found was limited. Therefore, the population rapidly converged to the (reduced) set of valid pipelines randomly generated in the initial population. 
	
	Our proposal in this paper is to define the IMs as a special kind of operator within the search space of the GP algorithm. This is performed via the introduction of a conceptual constraint in the algorithm. The restriction forces all the solutions to contain a single IM, and it can only be placed at the very beginning of the pipeline. This way, only pipelines that firstly solve the data incompleteness issue can be generated, and therefore the subsequent processes get imputed data as input, which they can handle. Additionally, we avoid introducing an imputer anywhere else in the pipeline, which would produce a sterile process, since the data has already been imputed in the first phase.
	
	Furthermore, the same idea implemented in this paper could be extended to introduce different types of ML operators to pipelines, considering restrictions similar to the ones regarded in this work.

	
	
	\section{Background}
	
	\subsection{Machine learning pipelines and automated machine learning}
	
	Data usually needs to be preprocessed before it can be used to build a ML model (i.e., a supervised classifier). Generally, these ML procedures can be organized and applied to the data in a sequential order, which suggests an obvious structure, the Pipeline. A pipeline is essentially a sequence of machine learning processes (usually various preprocessors, followed by a classification or a regression method) that are orderly applied to a piece of data, to produce a final model, ideally created from an optimal form of that data. 
	Sklearn \cite{pedregosa_scikit-learn:_2011}, a popular ML software implemented in Python, has its own representation of pipelines, with a homonym class. An example of an object of this class follows:
	
	\begin{lstlisting}
	Pipeline(steps=[((`fastica',FastICA(algorithm=`parallel', 
	fun=`logcosh', fun_args=None, max_iter=200, 
	n_components=None, random_state=None, tol=1.0, 
	w_init=None, whiten=True)), (`kneighborsclassifier',
	KNeighborsClassifier(algorithm=`auto', leaf_size=30,
	metric=`minkowski', metric_params=None, n_jobs=1, 
	n_neighbors=45, p=1, weights=`distance'))])
	\end{lstlisting}
	
	We can see how a piece of data processed by this pipeline would first be treated by the \textit{FastICA} preprocessing method, before being used by a 45-nearest-neighbor classifier algorithm.
	
	The approach used in this paper searches the space of pipelines that can solve a given ML problem using GP.
	
	\subsection{Grammar Guided Genetic Programming}
	
	GP was introduced in the early 90's by Koza \cite{koza_genetic_1990} as a paradigm of automatically ``creating" computing solutions for any class of problems. Instead of having a human-developed program, this concept proposes to seek for a solution in a search space where all possible programs can be found. This idea implements an evolutionary algorithm \cite{whitley_genetic_1994} to perform such exploration, considering evaluable programs as individuals (shaped as trees), and contemplating typical genetic operators such as crossover and mutation. GP has proven to be a good alternative to human knowledge when it comes to the automatic generation of programs, as it has sometimes produced programs as good as human-made versions, or has even improved classical program developments and implementation processes \cite{koza_annual_2017,banzhaf_genetic_1998,koza_genetic_1992}.
	
	One useful specification of the generic GP paradigm is the introduction of a grammar that each individual in all generations must adhere to, which is known as Grammar Guided GP (GGGP). This feature was already (implicitly) present in Koza's first book \cite{mckay_grammar-based_2010}.
	
	The grammar characteristic forces all the individuals to have a certain \textit{shape}, which can be very convenient in some contexts. GGGP works by defining classes, and assigning certain input and output parameters to all operators, so that only an operator with a certain class \textit{x} as an output can be placed as a child node (in the tree structure) of another operator with that same \textit{x} class as an input. A very detailed description of this solution, accompanied by graphical representations can be found in \cite{mckay_grammar-based_2010}.
	
	In addition to the benefit of shaping the individuals as pleased, this technique has also another profitable characteristic; the search space shrinking. Due to the fact that some combinations of operators are not accepted by the grammar, the amount of valid and, therefore, susceptible of being generated individuals decreases, resulting in a more efficient search.
	
	A straightforward way of implementing these grammars is by using \textit{strongly typed} GP \cite{montana_strongly_1995}.
	
	\subsection{TPOT, a platform for automatic parameter selection}
	
	TPOT (Tree-based Pipeline Optimization) \cite{Olson_applications_2016} exploits GP to evolve machine learning pipelines for regression and classification problems. The program implements a bi-objective GP algorithm \cite{barlow_incremental_2004,bleuler_multiobjective_2001} where pipelines of the popular sklearn  library are evolved. 
	
	TPOT selects a subset of the ML algorithms available in sklearn, and defines them as GP primitives. Primitives are organized in a tree structure to form individuals. These trees are the structures evolved aiming to obtain the optimal combination of processes.
	
	The component running TPOT is the DEAP (Distributed Evolutionary Algorithms in Python) library \cite{fortin_deap:_2012}. This package provides TPOT the necessary methods to implement genetic programming over the sklearn-based pipelines. DEAP is basically a framework that simplifies the task of creating fast evolutionary prototypes, which perfectly suits TPOT. The contribution of DEAP on this specific task of serving TPOT is explained in the following paragraphs.
	
	\begin{figure}
		\begin{center} 
			\scalebox{0.3}{\includegraphics{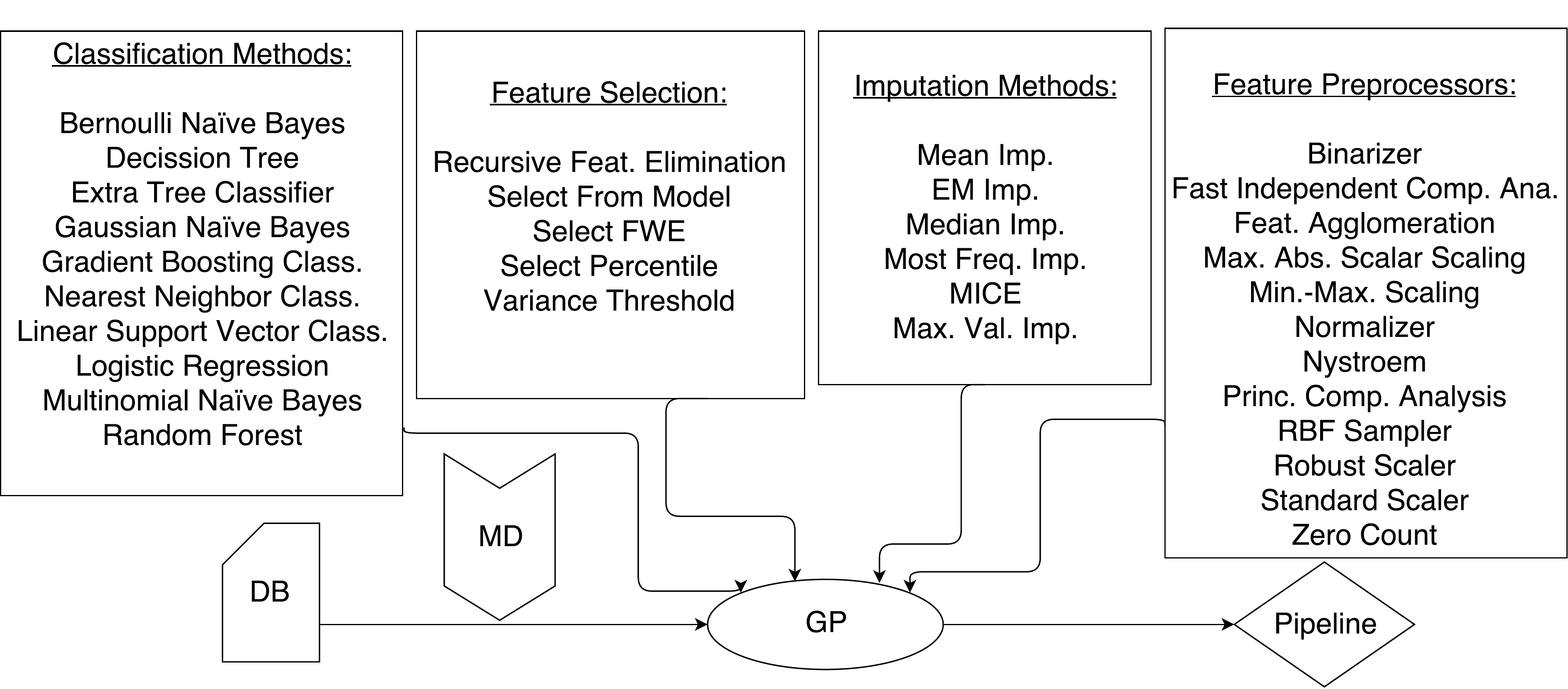}} 
			\caption{TPOT Workflow} 
			\label{fig:IMPUTATION} 
		\end{center} 
	\end{figure}
	
	Figure \ref{fig:IMPUTATION} shows the components underlying imputation-enhanced TPOT. Note that, in this work, we are focused in the supervised classification problem, but TPOT also implements regression methods. Though the whole work only references to the classification algorithms are made, but the imputation introduction is equally valid for regressors. 
    
    As it can be seen, all the components are fed to the GP module, which searches for an optimal pipeline to treat the data externally provided. The algorithm initialization phase consists of the creation of a population, which are a set of randomly generated-tree shaped sklearn pipelines. Each one of these pipelines are treated as individuals along the GP process. The individuals are evaluated regarding the accuracy they were able to obtain via balanced 3-fold cross-validation, and the amount of processes they encompass. This means that GP addresses a bi-objective optimization problem and solves this it by finding a Pareto front approximation. During the search, the top trees are selected to form the next generation, and new solutions are created using crossover and mutation operations. A \textit{hall of fame} is also updated with the trees in the current Pareto set. Once the new generation is created, the whole process is repeated, until an iteration limit criterion is reached.
	
	The multi-objective implementation of TPOT responds to the necessary length constraint set to the pipelines. Without this restriction, the software could lead to the production of pipelines composed by a very long chain of preprocessors. This would result in some pipelines failing to pass the time limit for an individual evaluation, therefore discarding  potentially \textit{good} solutions.
	
	However, this characteristic has a drawback. The final result is not a single individual, as a single-objective algorithm would produce, but a list of non-dominated individuals. Therefore, a final selection from the \textit{hall of fame} needs to be done. For this task, the individual with the best accuracy is selected, thus ignoring the simplicity objective this last time.
	
	TPOT has already been compared to various common selections of novice sklearn users \cite{elizalde_experiments_2016,olson_tpot:_2016,Olson_applications_2016}. TPOT regularly obtains results as good as other simpler search choices and hardly ever gets beaten.

	\subsection{Grammar in TPOT}
	
	The grammar TPOT uses in its current version consists on assigning specific input and output classes to the primitives (or sklearn operators). In this manner, some certain primitives will only be followed (or preceded) by some other certain primitives. 
	
	Currently, the grammar used in the design of TPOT \textit{only} considers two types of primitives; data preprocessing procedures (feature transformation and selection), and classifiers. The set the primitives belong to are determined by the type of a specific parameter they take as input, and the type of the result they produce. Leaving the parameters specific for each function (i.e., the number of neighbors for k-NN, or the kernel for a SVM) aside, the data preprocessors take as input a \texttt{ndarray} (from the Python \textit{numpy} library) parameter, the data we are willing to use to create a classifier. These operators produce another \textit{ndarray}, which contains the transformed version of the data. Classifiers exclusively accept an \texttt{ndarray} as their input (again, function-specific parameters aside), but this kind of algorithms can produce a different class, created specifically for TPOT, \texttt{Output\_array}. 
    
This system allows the creation of a grammar to structure solutions, but it does not affect their functionality. For example, the \textit{Output\_array} does not define nor contain any kind of new data structure, it is only used to differentiate the classifiers from the preprocessors and contains a regular \textit{ndarray} with the predictions made by the pipeline. The class utilized in the grammar, \textit{ndarray}, is a natural selection, since the pipeline actually collects a \textit{ndarray}, and the transformations also produce this same class.

Additionally, the overall individual is required to produce an \texttt{Output\_array}, therefore, only the primitives producing \texttt{Output\_array}s as result (only the classifiers) are allowed to be placed in the last step of the pipeline.
	
Note that, since the preprocessors both take and produce \texttt{ndarray}s, they can be stacked one after the other. Also, TPOT implements a stacking estimator method, and therefore classifiers can also produce \texttt{ndarray}s, and thus be placed as preprocessors.
	
\begin{figure}
		\begin{center} 
			\includegraphics[width=\textwidth]{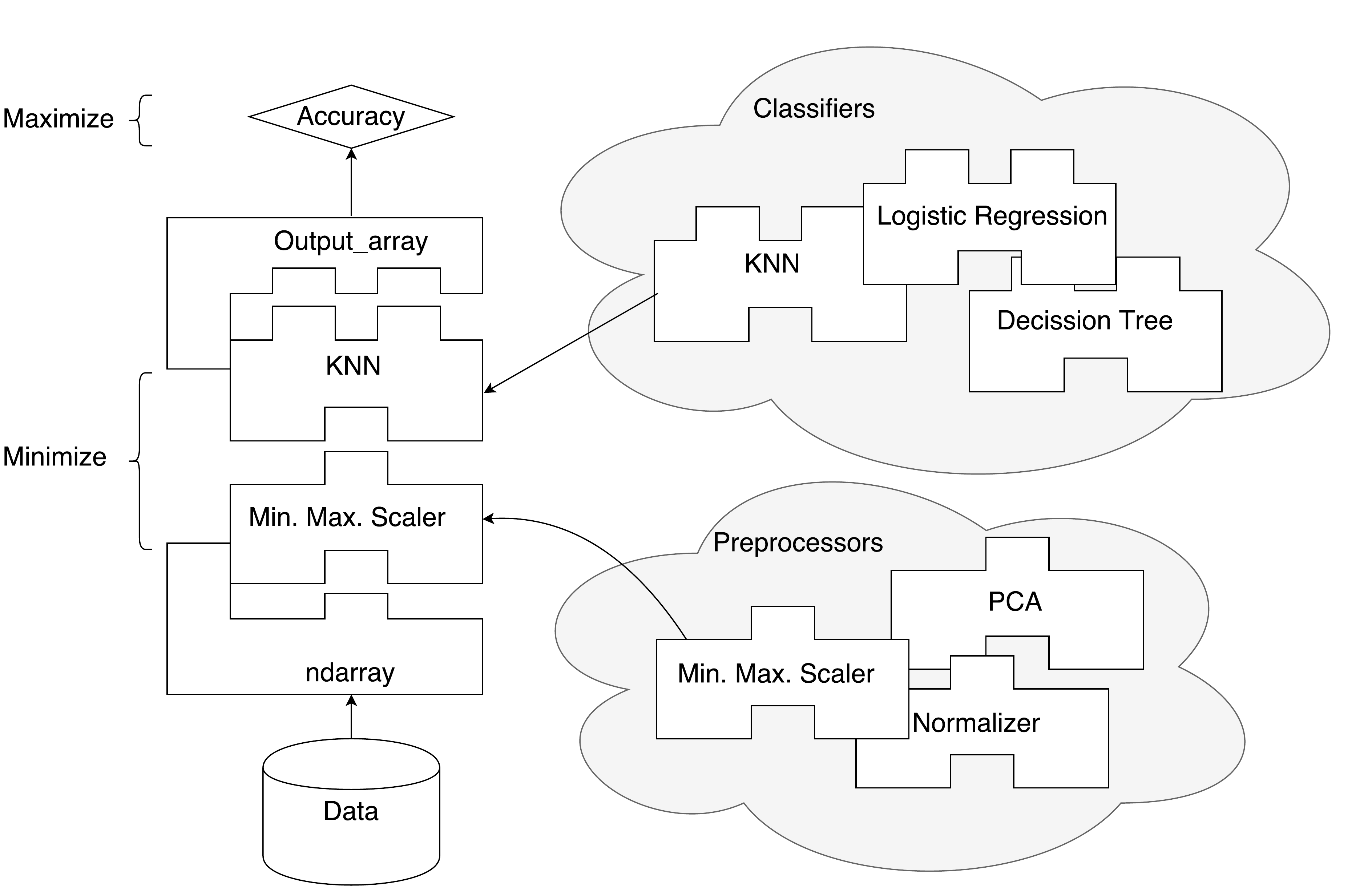}
			\caption{Graphical representation of the grammar used in TPOT. In this figure, the \textit{ndarray} python class in \textit{numpy} is represented by one spike, while the \textit{Output\_array} class created by the authors of TPOT is represented with two spikes.}
			\label{fig:GrammarSimple} 
		\end{center} 
	\end{figure}

	In Figure~\ref{fig:GrammarSimple} a simple graphical representation of the grammar can be found. In the figure we can see how only classifiers can produce the output required by TPOT, while preprocessors could be stacked one behind another endlessly. The length of a sequence of stacked preprocessors is however restricted also by the secondary \textit{pipeline shortness} objective in TPOT. Note that classifiers can also be part of the preprocessor section, but, in this figure, they have been omitted for simplicity.
	
	Additionally, Table \ref{tabl:origGram} shows a high level abstraction interpretation of the grammar ruling TPOT.
	
	\begin{table}
		\begin{center}
			\begin{tabular}{c}
				Original TPOT grammar \\\hline
				\textbf{CLAS} $\rightarrow$ \textit{sc} \textbf{PREP} \textit{sc\_args} \\
				\textbf{PREP} $\rightarrow$ \textit{prep} \textbf{PREP} \textit{prep\_args} $|$ \textit{mat}\\
			\end{tabular}
		\end{center}
		\caption{Deductible grammar in the original TPOT. The lower case symbols denote terminals, while the words in capitals represent grammar components. \textit{sc} refers to a classification algorithm from those available in TPOT. \textit{sc\_args} contains the necessary parameters for the previous classification algorithm. \textit{prep} represents a preprocessing algorithm of those present in TPOT, while \textit{prep\_args} stand for its adequate parameters. Finally, \textit{mat} is equivalent to the DB matrix being used to create the model. The maximum depth of the tree is fixed in the initial generation, to avoid infinite loops.}
		\label{tabl:origGram}
	\end{table}
	
	Introducing and exploiting a grammar is a very effective way of forcing the evolutionary algorithm to create pipelines of the desired shape (either via generation, crossover, or mutation), and also enables a developer to add another type of operator, or change the order these operators can be arranged.
	
	\section{Related Work}
	
	Some works that take advantage of the strongly typed GP to introduce a grammar to a GP problem have been developed, and the following paragraphs cover two interesting examples.
	
	Shan et al. \cite{shan_software_2002} used a grammatically constrained GP method to generate a classifier. The classifier was tested in a DB which contained records describing diverse software projects, with multiple relevant features, such as, the size of the team working on it, amount of defects, or hours of work invested, among others. The goal was to, given certain characteristics, predict the optimal values for other variables, as, for example, the language it should be implemented in. The approach was tested against linear and log-log regressions, on five separate runs. Five different measurements were used to compare performances. The GP approach obtained the best results by a long margin considering one of the measures (MSE, the one being minimized by the learners), while the differences considering the other four metrics showed no determinant differences.
	
	A similar work-flow was proposed by Ngan et al. in \cite{ngan_using_1998}. The authors proposed a grammar driven GP approach to induce rules from data and applied it to two real medical databases. One of the datasets produced rules that went against the common knowledge and what could be expected, which made the authors conclude that the approach is valid to uncover latent relations between variables.
	
    Note that these approaches use GGGP to evolve learners. The approach described in this paper exploits GGGP to develop a \textit{metalearner}.
	\section{Introducing imputation methods in TPOT:  penalty-based approach}
	
	\subsection{Previous approach}
	
	A first attempt to introduce IMs to TPOT was performed in \cite{garciarena_evolving_2017}. This attempt was simple and straightforward, and the results it produced were considerably biased towards the initial generation. The key characteristics of the approach compared to the regular TPOT are listed below:
	
	\begin{itemize}
		\item IMs were introduced into TPOT as regular preprocessors.
		\item Pipelines without or with non-imputing preprocessors could be generated, being therefore invalid.
		\item High chance of generating invalid pipelines, due to the large set of preprocessors.
	\end{itemize}
	
	The last characteristic caused that the invalid pipelines were assigned a negative fitness function, therefore not being selectable for the creation of the subsequent generation. Consequently, the final result was very biased towards the valid pipelines randomly constructed in the first generation, as results rapidly converged after few generations.

	The deductible grammar from this approach is similar to the one in the previous section, and can be found in Table \ref{tabl:secGram}.
	
	\begin{table}
		\begin{center}
			\begin{tabular}{c}
				Original TPOT grammar (with IMs added to operators set) \\\hline
				\textbf{CLAS} $\rightarrow$ \textit{sc} \textbf{PREP} \textit{sc\_args} \\
				\textbf{PREP} $\rightarrow$ \textit{prep} \textbf{PREP} \textit{prep\_args} $|$ \textit{imp} \textbf{PREP} \textit{imp\_args} $|$ \textit{mat}\\
			\end{tabular}
		\end{center}
		\caption{Deductible grammar in the original TPOT. The lower case symbols denote terminals, while the words in capitals represent grammar components. \textit{sc} refers to a classification algorithm from those available in TPOT. \textit{sc\_args} contain the necessary parameters for the previous classification algorithm. \textit{prep} represents a preprocessing algorithm of those present in TPOT, while \textit{prep\_args} stand for its adequate parameters. Analogously, \textit{imp} represents an IM, and \textit{imp\_args} are its parameters. Finally, \textit{mat} is equivalent to the DB matrix being used to create the model. The maximum depth of the tree is fixed in the initial generation, to avoid infinite loops}
		\label{tabl:secGram}
	\end{table}

	\subsection{Changing the TPOT grammar}
	
	In case we would like to define a new type of operator in the sklearn pipeline, or we would like to give a subset of processes a special condition, modifying the grammar would possibly be the most effective way to achieve the goal. This scenario can be faced because of multiple reasons. For example, if we want to significantly reduce the elapsed time while running the processes in a pipeline, forcing the structure to contain a variable selector, which reduces the cardinality of the set of features, would probably reduce the time consumed by the classifying algorithms. Also, if we had a database with various features with diverse scalings, we may want to force a preprocessing method that normalizes the features in order to avoid giving more weight to large scale variables rather than smaller ones in classifiers as k-NN.
	
	In such cases in which we need to \textit{force} some primitives into our structure, the most sophisticated manner to reach this goal is to declare a new class that is key in the structure, and grant this key to the processes we are interested in.
	
	\section{Introducing imputation methods in TPOT}
	
	To prove the effectiveness of the GGGP method presented in this paper, we have applied it to the specific case of imputation. In the version this work is based on (v0.8.3\footnote{https://github.com/rhiever/tpot/tree/995f8ccadbf395819b0b10223f7e7db2f0e812e3/tpot}), TPOT's regular way of dealing with missing data is by imputing data in an initial step (using median imputation), isolated from the GP procedure, then running TPOT normally. However, this strategy does not contemplate interactions between IMs, preprocessors and classification algorithms. For that reason, in order to incorporate imputation to the pipelines evolved by TPOT in a more sensible and efficient way, our new approach introduces a modification to the data structures and algorithm used by the software. The main characteristics of this new approach follow:
	
	\begin{itemize}
		\item Modification of the abstract grammar ruling TPOT using a new class. Introduction of new element, \emph{imputer}.
		\begin{itemize}
			\item Introduction of a new class into the strongly typed GP, \textit{imputed\_matrix}.
		\end{itemize}
		\item All individuals in all populations are forced to be valid.
		\item Similar behavior to regular TPOT, but with different IMs included in the pipeline search space.
		\item Force an imputer in the pipeline (while being created in the first generation) in case that only one more operator can be introduced before reaching the length limit.
	\end{itemize}
	
\begin{figure}
		\begin{center} 
			\includegraphics[width=\textwidth]{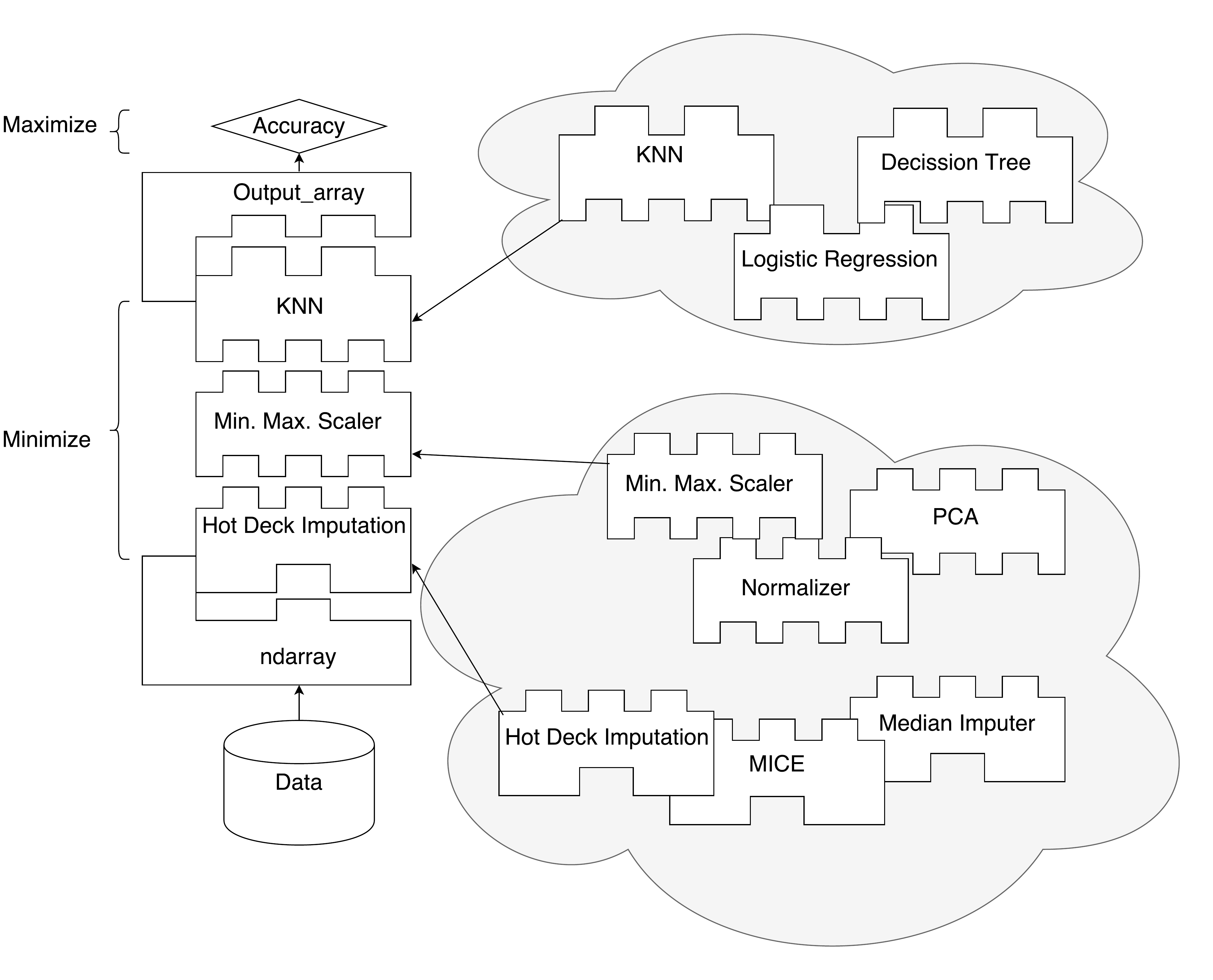}
			\caption{Graphical representation of the grammar used in the enhanced TPOT. As in the previous Figure \ref{fig:GrammarSimple}, the \textit{ndarray} class in \textit{numpy} is represented by one spike, while the \textit{Output\_array} class created by the authors of TPOT is represented with two spikes. In this case, the \textit{imputed\_matrix} has been added, and represented by three spikes. We can see how the IMs are indispensable to match the overall shape of the structure.}
			\label{fig:GrammarImp} 
		\end{center} 
	\end{figure}

	The main goal of the introduction of the abstract structure is to force all pipelines to be valid. This requires an imputer in the first position, followed by any combination that could be produced by the regular TPOT. To fulfill this task, the first step was to create a new class, \textit{imputed\_matrix}. Then, we made the production of this class exclusive to the imputing preprocessors. Finally, we set the input parameter of the rest of the operators in the search space to \textit{imputed\_matrix}. This way, all the processes present in the original TPOT could only take as input what only the imputers produce as output, therefore making the latter indispensable. As before, this can also be graphically interpreted, as in Figure~\ref{fig:GrammarImp}.
	
	Additionally, we had to give TPOT a little reminder when generating new aleatory pipelines. The length of the pipelines is randomly chosen before beginning to construct them, and this random number is chosen from a certain range. Even though the grammar has been fixed, the pipeline generator could produce a pipeline that does not fit the grammar, since it could place non-imputing preprocessors until the depth limit is reached, therefore producing an invalid structure. 
	
	The interpretable grammar resulting from the introduction of these new elements can be found in Table \ref{tabl:thirdGram}.
	
	\begin{table}
		\begin{center}
			\begin{tabular}{c}
				Modified TPOT grammar \\\hline
				\textbf{CLAS} $\rightarrow$ \textit{sc} \textbf{PREP} \textit{sc\_args} \\
				\textbf{PREP} $\rightarrow$ \textit{prep} \textbf{PREP} \textit{prep\_args} $|$ \textbf{IMP}\\
				\textbf{IMP} $\rightarrow$ \textit{imp} \textit{mat} \textit{imp\_args}
			\end{tabular}
		\end{center}
		\caption{Deductible grammar in the original TPOT. The quoted symbols denote terminals. The meaning of the parameters are as described in Tables \ref{tabl:origGram} and \ref{tabl:secGram}.}
		\label{tabl:thirdGram}
	\end{table}
	
	\section{Comparative experiments}
	
	To test the validity of our approach, we have designed a set of experiments that compare the first approach we presented \cite{garciarena_evolving_2017}, and the new technique proposed in this work.
    
    The first component of the experiments is the selection of the DB benchmark. We have selected a subset of 9 DBs of those available in \cite{olson_pmlb:penn_2016}. For the selection of the datasets we have taken into consideration different aspects, such as how difficult is for TPOT to find a good solution, or the amount of observations and features.
    
Table \ref{tab:DATASETS} contains a quantitative description of the selected DBs. The first three columns show the amount of variables of each type present in each dataset. The next two columns describe the total size of the data. The following one shows the index by which the DBs are identified from now on in this work. Finally, the last two columns describe the classes of the observations in each DB. The first of them ($n$) contains the number of classes, while the last one ($H_n$) shows the normalized entropy, computed as:

\begin{center}
	$H_n(p)=-\sum\limits_{i}\frac{p_i \cdot log_b(p_i)}{log_b(n)}$,
\end{center}

where $p_i$ denotes the probability of class $i$, $b$ represents the logarithm base ($e$ in this case). $H_n$ results in a number between $0$ and $1$, and represents how balanced the data is. A value near $1$ is to be interpreted as a large uncertainty surrounding the class, while a value near $0$ means that it is much more likely to find observations of a certain class rather than the others. Since this benchmark contains a multi-class DB, this coefficient is normalized by the amount of classes present. 

\begin{table}[ht]

	\begin{center}
		{$ \begin{array}{|c|c|c|c|c|c|c|c|c|} \hline 
			\text{Database}  	& \text{Disc.} & \text{Bin.} & \text{Cont.} & \text{N.Feats.} & \text{N.Cases} & \text{Ind.} & \text{n} & \text{$H_n$} \\ \hline 
			\text{AIDS}      	& 2  & 0  & 2 & 4  & 50  & 1 & 2 & 1    \\\hline 
			\text{Asbestos}  	& 1  & 1  & 1 & 3  & 83  & 2 & 2 & 0.99 \\\hline 
			\text{Bankruptcy}  	& 1  & 0  & 5 & 6  & 50  & 3 & 2 & 1    \\\hline 
			\text{Boxing1}    	& 2  & 1  & 0 & 3  & 120 & 4 & 2 & 0.93 \\\hline
			\text{Boxing2}    	& 2  & 1  & 0 & 3  & 132 & 5 & 2 & 1    \\\hline 
			\text{Cyyoung8092}  & 1  & 2  & 7 & 10 & 97  & 6 & 2 & 0.81 \\\hline 
			\text{Fraud} 		& 1  & 10 & 0 & 11 & 42  & 7 & 2 & 0.89 \\\hline 
			\text{Australian}   & 4  & 4  & 6 & 14 & 690 & 8 & 2 & 0.99 \\\hline 
			\text{Auto}   		& 13 & 3  & 9 & 25 & 202 & 9 & 5 & 0.95 \\\hline 
			\end{array}$}
            \vspace{.5cm}
		\caption{Numerical description of the DBs present in the benchmark.}
		\label{tab:DATASETS}
	\end{center}
\end{table}

The values in the $H_n$ column indicate that Adult, Cyyoung8092, and Fraud datasets are the only considerably unbalanced DBs.

To introduce MD in the DBs, we can find three or four different mechanisms in the literature, depending on the referred work \cite{batista_analysis_2003,garciarena_extensive_2017,luengo_study_2010}. One of the most common selection for the MD introduction algorithm is the \textit{missing completely at random} (MCAR) configuration \cite{batista_analysis_2003,luengo_study_2010}, which assumes that there exist no interactions between the missing values, or the missing values and the present ones. Additionally, we have arbitrarily chosen to introduce 30\% of MD in each DB. Even though it is very common to find this amount (or even higher) of MD in experimental sections, this is a fairly elevated percentage. It was chosen aiming to give a more challenging task to TPOT, with the goal of obtaining more contrast between solutions in terms of accuracy, as robustness to high MD is also an important feature to be boosted.

The algorithm used for introducing MCAR MD type is the one described in \cite{garciarena_extensive_2017}, and Algorithm \ref{alg:MCAR}.

\begin{algorithm}[ht]{
		\textbf{Input:}  \\data: Database\\mdp: MD percentage\\
		\textbf{Output:} Database with mdp$\%$ generated MD\\
		\Begin{
			x = numObservations(data)\\
			y = numVariables(data)
			
			\For{i $\in$ [0, x $\cdot$ y $\cdot$ mdp//100]}{data[random([0,x]), random([0,y])] = ``NaN"}
			\KwRet (data)\\
		}
	}
	
	\caption{MCAR generating algorithm.}
	\label{alg:MCAR}
\end{algorithm}

The sequence of the process composing the experiment consisted on selecting each of the databases, and introducing MD in them as described in Algorithm \ref{alg:MCAR}. Then, each incomplete DB is subjected to the imputation-enhanced TPOT process. The parameters selected for the TPOT runs were the default ones (100 generations, and $0.9$ and $0.1$ for the crossover and mutation rates, respectively, and accuracy and 3-fold cross-validation for evaluation of the models developed), except for the population size, which is changed from 100 to 400. Due to the stochastic nature of the process (both in the initial generation and the subsequent generational derivations). This process was run 30 times in each DB, to minimize the effect of the randomness component.
	
\subsection{Experiment results} \label{sec:results}
    
\begin{figure}
	\centering
	\includegraphics[width=\textwidth]{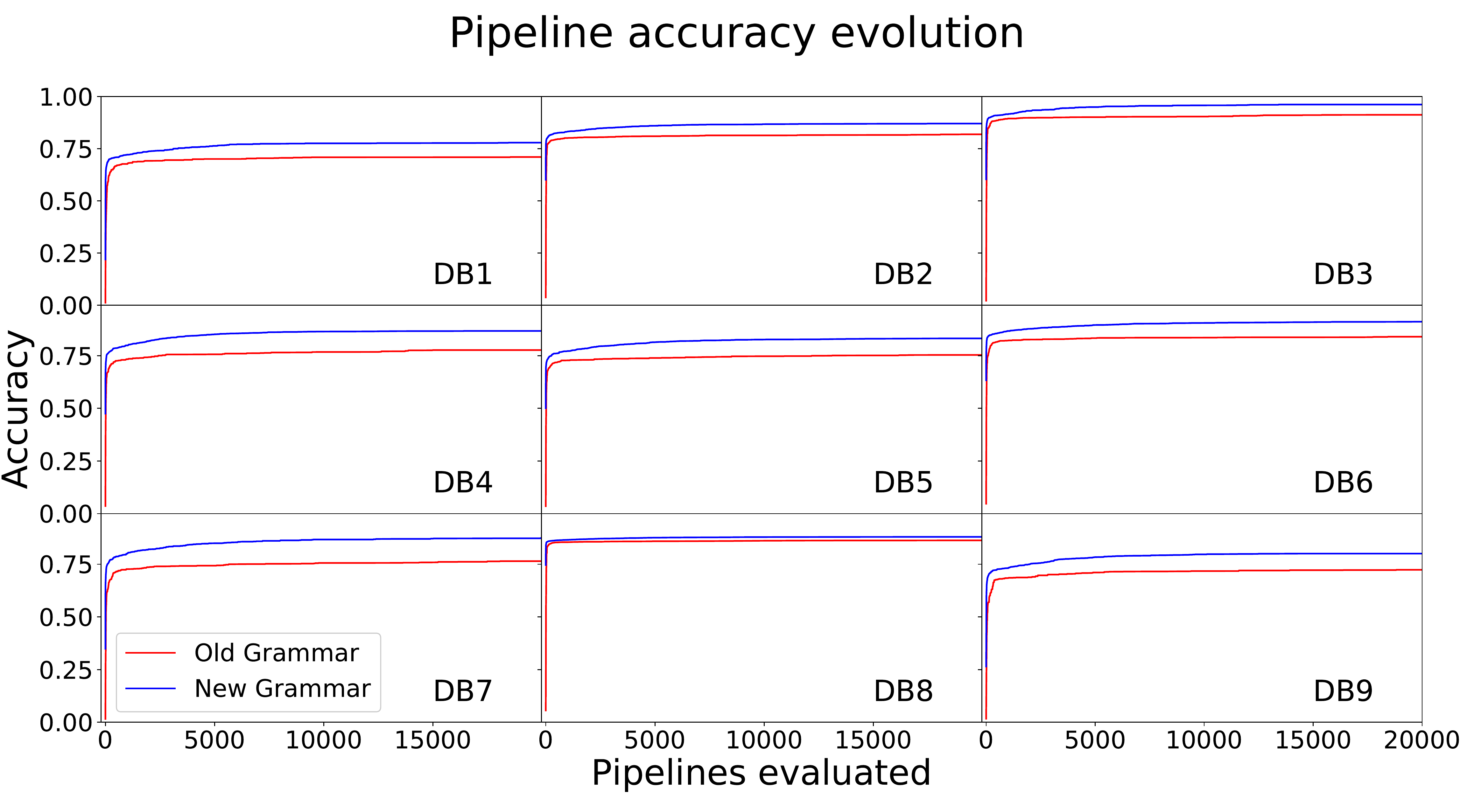}
	\caption{Evolution of the best pipeline found after each pipeline evaluation. To generate these curves, all the evaluations were separated by the DB they were generated from, and then averaged. 'y' axis shows the accuracy reached at each point, while 'x' axis shows how many pipelines have been evaluated. The best pipeline recorded was updated after each evaluation.}
    \label{fig:evols}
\end{figure}

From these 30 runs, the best pipeline found up until each moment is recorded. Figure \ref{fig:evols} shows the average values obtained by all 30 executions, discriminated by the DB it was generated from and whether the TPOT being run had the enhanced grammatic. The $y$ axis shows the mean accuracy, while the $x$ shows the number of pipelines evaluated in each run. 

Since the population size was $400$, and $100$ generations were run, it would have been expected to obtain a total of $40,000$ pipelines. However, TPOT, for runtime optimization, stores all the pipelines that have been evaluated, which prevents from evaluating the same pipeline twice, therefore reducing the time elapsed during the execution. For this reason, $40,000$ only stands as a upper bound of the evaluable pipelines during one run. The number of pipelines evaluated varies from one execution to another. We have decided to show the results until $20,000$ evaluations, a point by which the algorithm has almost converged, and close to the total number of evaluations, in most runs.

As it was to be expected, the results for each DB differ, as each DB has its own level of difficulty in terms of the accuracy that models built from each dataset can reach when performing a cross-validation evaluation. However a common pattern can be found in all of them. Due to the high probability of the old-grammar-equipped TPOT of generating invalid pipelines from scratch (for the first generation), the lines in Figure \ref{fig:evols} generated from this version of the software starts in a lower range than the one with the improved grammatic.

During the rest of the run, even though the evolution line follows a similar shape, it is the improved TPOT which consistently had better results. However, due to the high number of invalid pipelines that the previous enhanced TPOT does not evaluate (and therefore, virtually no time was invested in evaluating them), the runtime of that version was considerably lower.
    
	\section{Conclusions and future work}
	
	In this work a new approach to incorporate imputation methods to TPOT has been proposed. This approach exploits the grammar that is implicitly used by the strongly typed GP implemented in TPOT. The modified grammar results in an enhanced TPOT with a very similar methodology when it comes to the solution generation, but with various ways of dealing with incomplete data, at cost of a relatively small search space increase.
	
	We have compared our approach to the proposal of IM addition to TPOT made in \cite{garciarena_evolving_2017}. We have used 9 databases, and a 30 \% of missingness, introduced following a MCAR pattern. The results show that the evolution of pipelines is slightly better in the TPOT with the newly introduced grammatic compared to the approach presented in \cite{garciarena_evolving_2017}. The introduced method has an undesired side effect in the increase in the runtime of TPOT due to the increased number of pipelines to be evaluated.
	
	As future work, this approach leaves a door open to the introduction of new types of operators into TPOT. Following our methodology, investing little effort, new substructures of the existing elements could be forced into the sklearn pipelines, and new processes could be added, similarly to the way we have implemented the inclusion of imputation methods. One way to exploit this feature could involve an automated change of pipeline shape. As an example, when fitting the model, we could detect that the provided data is too large to be processed on its current state, and therefore needs to be processed by a dimensionality reduction process.
	
	We find another example when we face large-scale data. It is known that the support vector machine implementations in sklearn tends to take large amount of time to create a model from such data. To solve this problem, we could distinguish between two types of classifiers, the ones that can treat raw data and those which cannot, and force feature scalers in structures involving the latter.
	
	These are two problem examples a data scientist can face, and could eventually be automated if the grammars are correctly programmed.
	
	\section{Acknowledgments}

	This work was supported by IT-609-13 program (Basque Government) and TIN2016-78365-R (Spanish Ministry of Economy, Industry and Competitiveness) projects, while Unai Garciarena holds a predoctoral grant from the University of the Basque Country (ref. PIF16/238). We also thank Randal Olson for the recommendation of the DB benchmark from which the DBs in this work were gathered, and for his useful comments on the previous version of the report.
	
	\newpage
	
	\bibliographystyle{abbrv}

\begin{thebibliography}{10}
		
		\bibitem{banzhaf_genetic_1998}
		W.~Banzhaf, P.~Nordin, R.~E. Keller, and F.~D. Francone.
		\newblock {\em Genetic programming: an introduction}, volume~1.
		\newblock Morgan Kaufmann San Francisco, 1998.
		
		\bibitem{barlow_incremental_2004}
		G.~J. Barlow, C.~K. Oh, and E.~Grant.
		\newblock Incremental evolution of autonomous controllers for unmanned aerial
		vehicles using multi-objective genetic programming.
		\newblock In {\em Cybernetics and {Intelligent} {Systems}, 2004 {IEEE}
			{Conference} on}, volume~2, pages 689--694. IEEE, 2004.
		
		\bibitem{batista_analysis_2003}
		G.~E. Batista and M.~C. Monard.
		\newblock An analysis of four missing data treatment methods for supervised
		learning.
		\newblock {\em Applied Artificial Intelligence}, 17(5-6):519--533, 2003.
		
		\bibitem{bleuler_multiobjective_2001}
		S.~Bleuler, M.~Brack, L.~Thiele, and E.~Zitzler.
		\newblock Multiobjective genetic programming: {Reducing} bloat using {SPEA}2.
		\newblock In {\em Evolutionary {Computation}, 2001. {Proceedings} of the 2001
			{Congress} on}, volume~1, pages 536--543. IEEE, 2001.
		
		\bibitem{elizalde_experiments_2016}
		B.~Elizalde, A.~Kumar, A.~Shah, R.~Badlani, E.~Vincent, B.~Raj, and I.~Lane.
		\newblock Experiments on the {DCASE} {Challenge} 2016: {Acoustic} scene
		classification and sound event detection in real life recording.
		\newblock {\em arXiv preprint arXiv:1607.06706}, 2016.
		
		\bibitem{fortin_deap:_2012}
		F.-A. Fortin, F.-M.~D. Rainville, M.-A. Gardner, M.~Parizeau, and C.~Gagné.
		\newblock {DEAP}: {Evolutionary} {Algorithms} {Made} {Easy}.
		\newblock {\em Journal of Machine Learning Research}, 13:2171--2175, July 2012.
		
		\bibitem{garciarena_extensive_2017}
		U.~Garciarena and R.~Santana.
		\newblock An extensive analysis of the interaction between missing data types,
		imputation methods, and supervised classifiers.
		\newblock {\em Expert Systems with Applications}, 89:52--65, 2017.
		
		\bibitem{garciarena_evolving_2017}
		U.~Garciarena, R.~Santana, and A.~Mendiburu.
		\newblock Evolving imputation strategies for missing data in classification
		problems with {TPOT}.
		\newblock {\em arXiv:1706.01120 [cs, stat]}, June 2017.
		\newblock arXiv: 1706.01120.
		
		\bibitem{koza_annual_2017}
		J.~Koza.
		\newblock {\em Annual "{Humies}" {Awards} {For} {Human}-{Competitive}
			{Results}}.
		\newblock 2017.
		
		\bibitem{koza_genetic_1990}
		J.~R. Koza.
		\newblock {\em Genetic programming: {A} paradigm for genetically breeding
			populations of computer programs to solve problems}.
		\newblock Stanford University, Department of Computer Science, 1990.
		
		\bibitem{koza_genetic_1992}
		J.~R. Koza.
		\newblock {\em Genetic {Programming}: {On} the {Programming} of {Computers} by
			{Means} of {Natural} {Selection}}.
		\newblock The MIT Press, Cambridge, MA, 1992.
		
		\bibitem{luengo_study_2010}
		J.~Luengo, S.~García, and F.~Herrera.
		\newblock A study on the use of imputation methods for experimentation with
		{Radial} {Basis} {Function} {Network} classifiers handling missing attribute
		values: the good synergy between {RBFNs} and {EventCovering} method.
		\newblock {\em Neural Networks}, 23(3):406--418, 2010.
		
		\bibitem{mckay_grammar-based_2010}
		R.~I. Mckay, N.~X. Hoai, P.~A. Whigham, Y.~Shan, and M.~O’neill.
		\newblock Grammar-based genetic programming: a survey.
		\newblock {\em Genetic Programming and Evolvable Machines}, 11(3-4):365--396,
		2010.
		
		\bibitem{montana_strongly_1995}
		D.~J. Montana.
		\newblock Strongly typed genetic programming.
		\newblock {\em Evolutionary computation}, 3(2):199--230, 1995.
		
		\bibitem{ngan_using_1998}
		P.~S. Ngan, M.~L. Wong, K.~S. Leung, and J.~C. Cheng.
		\newblock Using grammar based genetic programming for data mining of medical
		knowledge.
		\newblock {\em Genetic Programming}, pages 254--259, 1998.
		
		\bibitem{olson_pmlb:penn_2016}
		R.~Olson and W.~L. Cava.
		\newblock {\em {PMLB}:{Penn} {Machine} {Learning} {Benchmarks}}.
		\newblock 2016.
		
		\bibitem{olson_tpot:_2016}
		R.~S. Olson and J.~H. Moore.
		\newblock {TPOT}: {A} {Tree}-based {Pipeline} {Optimization} {Tool} for
		{Automating} {Machine} {Learning}.
		\newblock In {\em Workshop on {Automatic} {Machine} {Learning}}, pages 66--74,
		2016.
		
		\bibitem{Olson_applications_2016}
		R.~S. Olson, R.~J. Urbanowicz, P.~C. Andrews, N.~A. Lavender, L.~C. Kidd, and
		J.~H. Moore.
		\newblock Applications of {Evolutionary} {Computation}: 19th {European}
		{Conference}, {EvoApplications} 2016, {Porto}, {Portugal}, {March} 30 –
		{April} 1, 2016, {Proceedings}, {Part} {I}.
		\newblock pages 123--137. Springer International Publishing, 2016.
		
		\bibitem{pedregosa_scikit-learn:_2011}
		F.~Pedregosa, G.~Varoquaux, A.~Gramfort, V.~Michel, B.~Thirion, O.~Grisel,
		M.~Blondel, P.~Prettenhofer, R.~Weiss, and V.~Dubourg.
		\newblock Scikit-learn: {Machine} learning in {Python}.
		\newblock {\em The Journal of Machine Learning Research}, 12:2825--2830, 2011.
		
		\bibitem{shan_software_2002}
		Y.~Shan, R.~I. McKay, C.~J. Lokan, and D.~L. Essam.
		\newblock Software project effort estimation using genetic programming.
		\newblock In {\em Communications, {Circuits} and {Systems} and {West} {Sino}
			{Expositions}, {IEEE} 2002 {International} {Conference} on}, volume~2, pages
		1108--1112. IEEE, 2002.
		
		\bibitem{whitley_genetic_1994}
		D.~Whitley.
		\newblock A genetic algorithm tutorial.
		\newblock {\em Statistics and computing}, 4(2):65--85, 1994.
		
	\end{thebibliography}

\end{document}